\documentclass[preprint,12pt]{elsarticle}
\usepackage[T1]{fontenc}
\usepackage[latin9]{inputenc}
\usepackage{appendix}
\usepackage{fancyhdr}
\usepackage{color}
\usepackage{babel}
\usepackage{amsmath}
\usepackage{bm}
\usepackage{amssymb}
\usepackage{graphicx}
\usepackage[a4paper]{geometry}
\usepackage{setspace}
\usepackage{enumitem}
\usepackage{algorithm}
\usepackage{algpseudocode} 
\usepackage{setspace}
\usepackage[bookmarks=true,bookmarksnumbered=false,bookmarksopen=false,
 breaklinks=false,pdfborder={0 0 0},pdfborderstyle={},backref=false,colorlinks=false]
 {hyperref}
\hypersetup{
 pdfauthor={}}
\usepackage{multirow}
\makeatletter

\usepackage{times}
\renewcommand{\@openbib@code}{\setlength{\itemsep}{-1pt}} 
\renewcommand{\subsectionmark}[1]{}

\usepackage[compact]{titlesec}
\titleformat{\section}{\LARGE \bfseries}{\thesection}{1em}{}
\titleformat{\subsection}{\large \bfseries}{\thesubsection}{1em}{}
\makeatother

\begin{document}
\global\long\def\d{\mathrm{d}}%
\global\long\def\defgradT{\mathbf{F}}%
\global\long\def\cT{\mathbf{C}}%
\global\long\def\stretch{\lambda}%
\global\long\def\stretchr{\lambda_{r}}%
\global\long\def\ang{\varphi}%
\global\long\def\torsion{T}%

\global\long\def\eone{\mathbf{\hat{e}}_{1}}%
\global\long\def\etwo{\mathbf{\hat{e}}_{2}}%
\global\long\def\ethree{\mathbf{\hat{e}}_{3}}%
\global\long\def\zhat{\hat{\mathbf{z}}}%
\global\long\def\rhat{\hat{\mathbf{r}}}%
\global\long\def\anghat{\hat{\boldsymbol{\theta}}}%

\global\long\def\secpiolaT{\mathbf{S}}%
\global\long\def\piolaT{\mathbf{P}}%
\global\long\def\stressT{\boldsymbol{\sigma}}%
\global\long\def\shear{\mu}%
\global\long\def\cona{\alpha_{2}}%
\global\long\def\conb{\alpha_{3}}%
\global\long\def\cong{\gamma}%
\global\long\def\piola{P}%
\global\long\def\stress{\sigma}%
\global\long\def\ro{r_{o}}%

\global\long\def\history{D}%
\global\long\def\SEDF{W}%
\global\long\def\pressure{p}%
\global\long\def\rateU{\dot{\lambda}}%
\global\long\def\rateT{\dot{\varphi}}%
\global\long\def\bulkm{\kappa}%

\begin{frontmatter}
\title{{\bf Physics Augmented Machine Learning Discovery of Composition-Dependent Constitutive Laws for 3D Printed Digital Materials}}

\author{Steven Yang$^{\dag}$}
 \address{{\it  Sibley School of Mechanical and Aerospace Engineering, Cornell University, Ithaca, NY 14850}}
 
\author{Michal Levin$^{\dag}$}
 \address{{\it  Department of Materials Science and Engineering, Technion  - Israel Institute of Technology, Haifa 3200003, Israel}}

 \author{Govinda Anantha Padmanabha}
 \address{{\it  Sibley School of Mechanical and Aerospace Engineering, Cornell University, Ithaca, NY 14850}}

\author{Miriam Borshevsky}
 \address{{\it  Department of Materials Science and Engineering, Technion  - Israel Institute of Technology, Haifa 3200003, Israel}}
 
 \author{Ohad Cohen}
 \address{{\it  Department of Materials Science and Engineering, Technion  - Israel Institute of Technology, Haifa 3200003, Israel}}
 
 \author{D. Thomas Seidl}
 \address{{\it Sandia National Laboratories, Albuquerque, NM 87123}}
 
 \author{Reese E. Jones}
 \address{{\it Sandia National Laboratories, Livermore, CA 94551}}
 
 \author{Nikolaos Bouklas$^*$}
 \address{{\it  Sibley School of Mechanical and Aerospace Engineering, Cornell University \& Center for Applied Mathematics, Ithaca, NY 14850\\
 Pasteur Labs, Brooklyn, NY 11205}}

  \author{Noy Cohen$^{**}$}
 \address{{\it  Department of Materials Science and Engineering, Technion  - Israel Institute of Technology, Haifa 3200003, Israel}}
 
\cortext[cor1]{Corresponding authors: nb589@cornell.edu}
\cortext[cor2]{noyco@technion.ac.il}
\cortext[equal]{$^{\dag}$ Equal contributions}

\begin{abstract}

Multi-material 3D printing, particularly through polymer jetting, enables the fabrication of digital materials by mixing distinct photopolymers at the micron scale within a single build to create a composite with tunable mechanical properties. This work presents an integrated experimental and computational investigation into the composition-dependent mechanical behavior of 3D printed digital materials. We experimentally characterize five formulations, combining soft and rigid UV-cured polymers under uniaxial tension and torsion across three strain and twist rates. The results reveal nonlinear and rate-dependent responses that strongly depend on composition. To model this behavior, we develop a physics-augmented neural network (PANN) that combines a partially input convex neural network (pICNN) for learning the composition-dependent hyperelastic strain energy function with a quasi-linear viscoelastic (QLV) formulation for time-dependent response.  The pICNN ensures convexity with respect to strain invariants while allowing non-convex dependence on composition. To enhance interpretability, we apply $L_0$ sparsification. For the time-dependent response, we introduce a multilayer perceptron (MLP) to predict viscoelastic relaxation parameters from composition. The proposed model accurately captures the nonlinear, rate-dependent behavior of 3D printed digital materials in both uniaxial tension and torsion, achieving high predictive accuracy for interpolated material compositions. This approach provides a scalable framework for automated, composition-aware constitutive model discovery for multi-material 3D printing. 

\end{abstract}
\end{frontmatter}
\section{Introduction}

3D printing and additive manufacturing enables the fabrication of complex geometries, composite structures, and architectured materials that are difficult or impossible to achieve with traditional manufacturing techniques. Initially dominated by thermoplastic polymers \citep{sharma&Shakti19MRI,jakus19book,gokh&etal17IJERT}, the field has expanded to include metals \citep{visser&Etal15AM,gibson18MT,ribe98CCEJ}, ceramics \citep{chen&etal19JECS,hwa&etal17COSSMS}, composites \citep{wang&etal17CPBE,comp&lewi14AM}, biological tissues \citep{sharma&Shakti19MRI,yan&etal18e,liu20MD}, and soft resins \citep{tee&etal20jom,lei&etal19MH}. Common 3D printing methods include fused deposition modeling (FDM), stereolithography (SLA), selective laser sintering (SLS), and polymer jetting \citep{gokh&etal17IJERT,kara&lin20COCE,patpatiya&etal22review,hanuhov2024design}. Among these, polymer jetting stands out for its ability to print intricate, multi-material structures with high spatial resolution in a single step. 

PolyJet, a commercial polymer jetting process developed by Stratasys Ltd., builds parts by jetting and UV-curing photopolymer droplets layer by layer. After each layer is cured, the build tray lowers along the Z-axis, allowing the next layer to be deposited. A multi-nozzle print head, with each nozzle dispensing a different photopolymer, enables precise geometries with spatially varying material composition and properties. Stratasys uses the term digital materials to describe the droplet-by-droplet combination of different polymers at the micron scale. Polymer jetting is distinct from traditional manufacturing methods because, instead of producing homogeneous blends, it creates composite structures in situ with microscale control over material distribution. This capability supports generating a range of soft rubber-like and stiff thermoplastic-like materials enabling fabrication of parts with diverse mechanical properties leveraged in medical devices \citep{shannon203dPAM,rachmiel17BJOMS,barone16IJIDeM,singh23Bioeng},
functional models for 4D-printing \citep{kuang&etal19AFM,wu&etal18CJPS,raviv&etal14SR},
and actuators \citep{slesarenko2018strategies,shi&etal20IEEEth,zhang&etal19AFM,levin&cohen23AMT}. This versatility has motivated experimental studies on individual and blended materials, including Tango Black, Agilus, Vero, and Digital ABS (RGD-515 and RGD-531) \citep{abay&Ghaj20AM,tee&etal20jom,sles&rudy18IJES,volpato&etal16IJMPT}. While these materials are generally viscoelastic, the constitutive behavior of Agilus-Digital ABS blends, particularly under torsion, remains largely unexplored despite its importance in many applications \citep{wine09MMS,aziz&Spinks20MH,singh&etal17JE,emuna&cohen21EML,xiao&etal21JMR,emuna&cohen21JAM,bazaev&cohen22IJSS,kole22CMT}.

The ability of polymer jetting to seamlessly blend photopolymers during fabrication creates a vast and continuously tunable space of material compositions and mechanical behaviors. Traditional constitutive modeling approaches, where separate models are fit for each composition, quickly become impractical in this setting. With advancements in machine learning, physics-augmented neural networks (PANNs) offer a promising approach to automated constitutive model discovery by embedding physical laws directly into the learning process~\cite{fuhg2024extreme,padmanabha2024condensed}. In material modeling, surrogate models developed by incorporating physical principles during training have proven to be more efficient than traditional data-driven models. Critically, the physics-augmented neural networks enables data-driven constitutive models trained on low-data with predictive capabilities to out of distribution data \cite{fuhg2024extreme,padmanabha2024condensed}. Recent applications in modeling hyperelastic \cite{klein_parametrized_2023,tac_data-driven_2022} and viscoelastic \cite{qin_physics-guided_2024,rosenkranz_viscoelasticty_2024} materials have demonstrated that PANNs effectively learn models that adhere to established thermodynamic and kinematic constraints. Notably, PANNs have been extended to composition-aware modeling, where material composition is treated as an input to the neural network model \cite{jadoon_inverse_2024}. These methods have been successfully applied across a range of material systems, including metals \cite{fuhg_machine-learning_2022}, polymers \cite{fuhg2024extreme,jailin_experimental_2024}, biomaterials \cite{tac_data-driven_2022}, and composites \cite{kalina_neural_2024}.

In this work, we focus on Agilus and its combination with Digital ABS, typically denoted as digital materials (DMs), and report the response of these materials under tension and torsion. We tested five compositions named A, DM-40, DM-50, DM-60 and DM-70, each differing by approximately 10 Shore A hardness units. To model this family of materials, we propose the following composition-aware PANN framework that employs a partial input convex neural network (pICNN). The pICNN ensures that the learned strain energy density function satisfies the polyconvexity conditions required for hyperelastic constitutive modeling. Strain invariants are input into the convex branch of the network, while the composition ratio of Agilus to Digital ABS is fed into the non-convex branch. Instead of directly predicting stresses, our approach learns an intermediate scalar potential via the pICNN, from which the stress responses are derived. To improve interpretability and reduce model complexity, we incorporate $L_0$ sparsification~\cite{louizos2017learning}, which directly enforces parameter sparsity without penalizing weight magnitudes. To model the strain-rate and composition-dependent behavior of digital materials, we embed this framework within a quasi-linear viscoelastic (QLV) formulation, using a multilayer perceptron (MLP) to predict a composition-dependent relaxation coefficient. This ensures the viscoelastic response remains physically meaningful while maintaining model flexibility.

The paper is structured as follows: Section \ref{sec:Experimental} describes the experimental methods used to characterize the mechanical response of the digital materials. Section \ref{sec:Constitutive} presents the composition-aware constitutive modeling framework based on a physics-augmented neural network (PANN). In Section \ref{sec:Results-and-discussion}, we report the experimental findings, model training details, and model performance evaluation. Finally, Section \ref{sec:Conclusions} summarizes the key results and discusses future directions.

\section{Experimental Methods \label{sec:Experimental} }

In the following, \textcolor{black}{we characterize soft 3D printed
materials that are commercially available for the Stratasys Objet260
Connex3 printer. The materials we investigate are the rubber-like
Agilus and four combinations of Agilus and stiff Digital ABS (RGD-515
and RGD-531). The combinations are termed FLX98XX by Stratasys, where
XX denotes the Shore A hardness scale. For convenience, these materials
are referred to as digital materials (DMs) and are denoted by DM-XX.
For example, FLX9860 is a DM with a Shore A scale of 60 that is hereby
denoted DM-60. }In this work we focus on DM-40, DM-50, DM-60, and
DM-70. 
\textcolor{black}{We emphasize that higher Shore A index
materials are stiffer and contain larger fraction of Digital ABS. }

\subsection{\textcolor{black}{Tension experiment}}

\textcolor{black}{To measure the response under tension, dogbone-shaped specimens with a neck length {$60\,\mathrm{mm}$}, width {$15\,\mathrm{mm}$}, and thickness of {$2\,\mathrm{mm}$} were printed. All samples were printed along the same
direction to eliminate any possible influences of the printing-induced
orientation. Before testing, samples were left on the printing tray
at least 24 hours to allow post-printing curing. Next, the support
resin, SUP705 by Stratasys, was removed by high-pressure water cleaning
and left to dry for another 24 hours. To ensure reproducibility, three
samples of each material were printed and examined. }

\begin{figure}[ht]
\centering
\includegraphics[height=5cm]{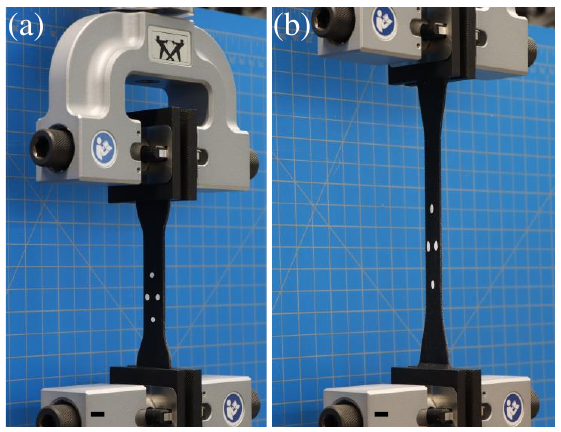}
\caption{Tension experimental set-up: DM-60 sample marked
with two longitudinal and two transverse marks, before (a) and during
(b) testing.}
\label{fig:Tension-exp}
\end{figure}

\textcolor{black}{The tension experiments were carried out on an Instron
5943 with an AVE 2 video-extensometer, which allows to measure the
longitudinal and the transverse strains in the middle of the sample.
The experimental set-up and a sample before and during testing are
shown in Fig. (\ref{fig:Tension-exp}). Using displacement controlled
mode, three displacement rates were applied to capture the viscoelastic
time-dependent response}: $\dot{u}=5,50,$ and $500\,\mathrm{mm/min}$.
These correspond to stretch rates of $\rateU=0.9\cdot10^{-3},0.9\cdot10^{-2},$
and $0.9\cdot10^{-1}\,\text{s}^{-1}$, where the stretch $\stretch\left(t\right)=\rateU t+1$
is the ratio between the deformed and the initial length of a line
element along the direction of the dog-bone. All samples were stretched
to rupture.

\subsubsection{\textcolor{black}{Poisson's ratio of digital materials}}

\textcolor{black}{Additionally, Poisson's ratio was computed from
the uniaxial extension tests under the slowest examined stretch rate $\rateU=0.9\cdot10^{-3}\,\text{s}^{-1}$
and at the elasto-linear region, under small stretches $\stretch<1.05$. }

\subsection{\textcolor{black}{Torsion experiment}}

To examine the response to torsion, cylinder
shaped samples {with length {$57\,\mathrm{mm}$} and rod diameter of \mbox{$10\,\mathrm{mm}$}} were printed.
Here again, the same protocol as in tension experiments was applied:
all samples were printed in the same orientation, then left on the
printing tray at least 24 hours. Afterwards, the samples were cleaned
from the support resin and left to dry for another 24 hours before
testing. Three samples of each formulation were examined.

\begin{figure}[ht]
\centering
\includegraphics[height=5cm]{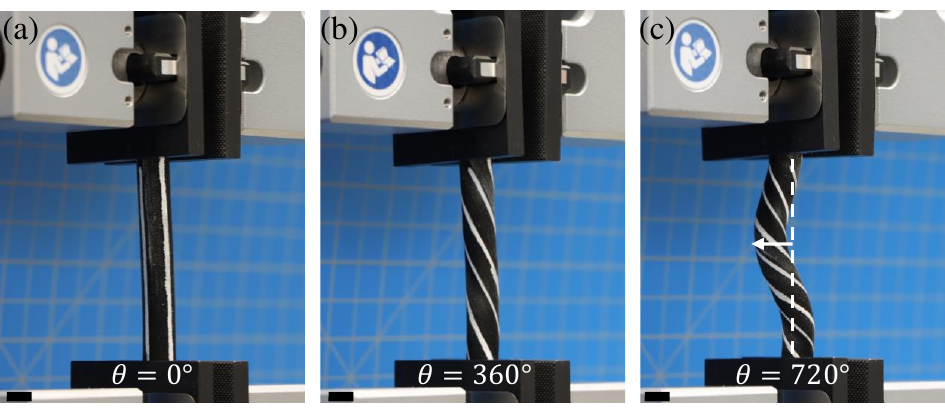}
\caption{Torsion experimental set-up: (a-c) A sample at twist angles of
$\protect\ang=0,\,360^{\circ},\,720^{\circ}$, respectively. The twist
is marked (solid lines), around the symmetry line (dashed line). At
$\protect\ang>360^{\circ}$ the buckling is marked with an arrow.
Scale bar: $1\,cm$.}
\label{fig:Evolution-of-cracks_torsion}
\end{figure}

\textcolor{black}{The torsion tests were conducted on Instron 5943
with a torsion add-on, under three different twist rates, specifically:
$\rateT=90,180,360\,\mathrm{\frac{deg}{min}}$. The experimental set-up
is presented in Fig. (\ref{fig:Evolution-of-cracks_torsion}a). In
this case, the twist depends linearly on the time: $\ang\left(t\right)=\rateT t$.
We note that the torsion results are shown for $\ang\leq360^{\circ}$
(Fig. (\ref{fig:Evolution-of-cracks_torsion}b)), as we wish to focus
on the elastic response. At higher twist angles micro-cracks start
to evolve and the rod buckles and deforms out-of-plane irreversibly,
as demonstrated in Fig. (\ref{fig:Evolution-of-cracks_torsion}c).
}

\section{Constitutive Modeling}\label{sec:Constitutive}

In this section, we introduce a physics-augmented neural network based on the input convex neural network (ICNN) framework introduced by \citet{amos2017input}, which has recently been applied to modeling hyperelastic materials by enforcing the convexity of the strain energy function with respect to strain invariants~\cite{fuhg2024extreme,padmanabha2024condensed}. Recently, \citet{jadoon2024inverse} extended the ICNN framework by introducing partially input convex neural networks (pICNNs), wherein the strain energy potential is enforced to be convex with respect to deformation measures while allowing
arbitrary non-convex dependence on design or material parameters. 

\begin{figure}[h]
    \centering
    \includegraphics[width=0.9\linewidth]{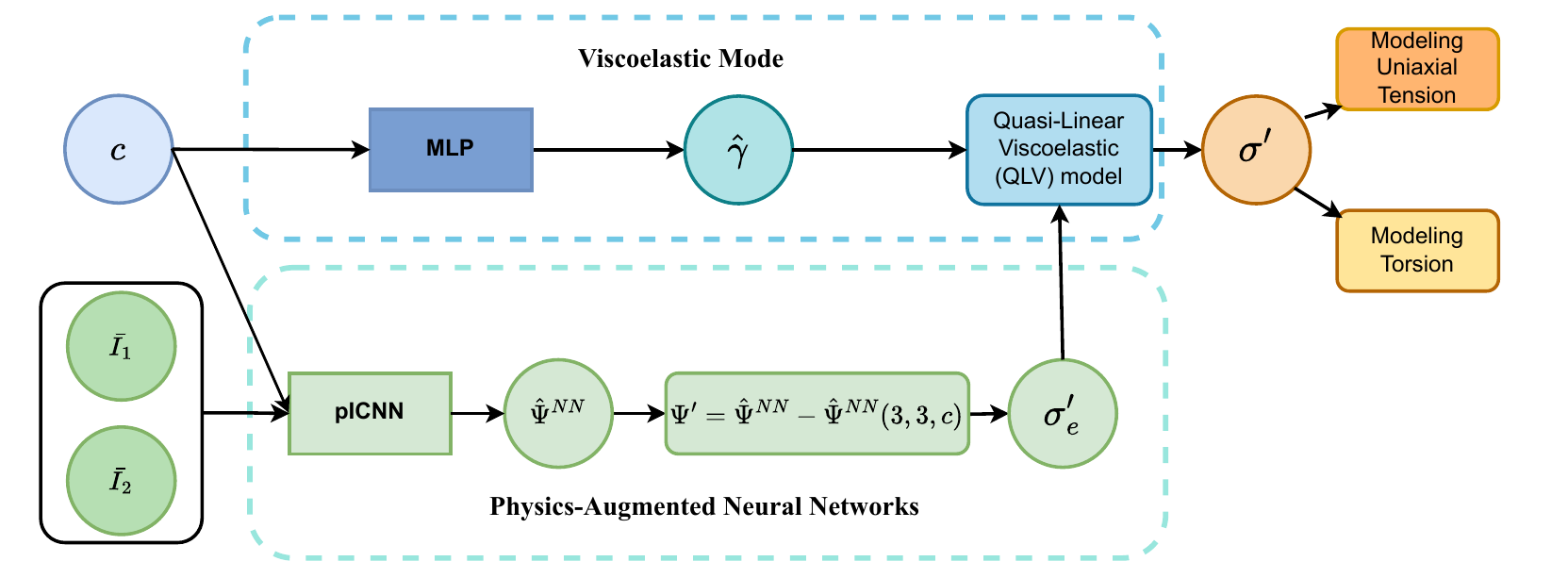}
    \caption{Flow chart of the constitutive modeling framework. Strain invariants and material composition (Agilus/Digital ABS ratio) are fed into convex and non-convex branches of the pICNN, respectively, to predict hyperelastic strain energy and instantaneous stress. Material composition also informs a separate MLP that outputs the viscoelastic parameter $\gamma$, which controls time-dependent stress via the quasi-linear viscoelastic model. The model predicts stress in tension or torque in torsion, depending on the deformation mode.}
    \label{fig:Model_flowchart}
\end{figure}

We extend the pICNN framework to model materials exhibiting both strain-rate dependence and composition-dependent behavior. The pICNN framework is employed to learn the instantaneous, rate-independent hyperelastic response, which is subsequently integrated into a quasi-linear viscoelastic (QLV) model to capture the complete time-dependent mechanical behavior. A schematic illustration of the proposed framework is presented in Figure~\ref{fig:Model_flowchart}. Specifically the convex part of the pICNN network takes strain invariants as input, while the non-convex branch receives the material composition information represented as the mixing ratio of Agilus and Digital ABS (Section \ref{sec:model_picnn}). The network outputs the instantaneous hyperelastic strain energy, from which the instantaneous stress is derived (Section \ref{sec:model_instantaneous_stress}). The material composition is also fed into a separate multilayer perceptron (MLP) that predicts the viscoelastic parameter $\gamma$, which controls the integration kernel in the quasi-linear viscoelastic (QLV) model (Section \ref{sec:model_qlv}). This setup allows ${\gamma}$ to vary with material composition, capturing differences in strain-rate sensitivity across mixtures. The QLV model is used to compute the time-dependent stress, which is then used to predict stress in uniaxial tension deformation data (Section \ref{sec:model_tension}) or integrated further to predict torque in torsion deformation (Section \ref{sec:model_torsion}).

\subsection{Physics-augmented neural network formulations}
\label{sec:method physics-aug NN}
In our work, for the physics-augmented neural network  (PANN) framework, an important aspect of physics-based enhancement in constitutive modeling involves the use of a partial input Convex Neural Network (pICNN). Convexity, in particular, plays a crucial role in ensuring that the learned strain energy density functions satisfy the polyconvexity condition required for hyperelastic materials. Rather than directly mapping strain invariants to stress outputs, our approach first maps the invariants to an intermediate scalar potential, from which the stress response is subsequently derived and incorporated into the loss function. Specifically, the final scalar-valued output representing the predicted strain energy potential is obtained through an ICNN.
\par
To improve interpretability and reduce model complexity, we incorporate $L_0$ sparsification~\cite{louizos2017learning}, which encourages a minimal set of active parameters by directly enforcing sparsity. Unlike approaches that penalize weight such as Lasso or Ridge regression, magnitudes $L_0$ regularization permits parameters to be set exactly to zero. This strategy preserves model flexibility while promoting compact and physically interpretable representations.

\subsubsection{Modeling of Instantaneous Response with PICNN}
\label{sec:model_instantaneous_stress}

The instantaneous stress response is modeled as hyperelastic, with material behavior governed by the strain energy density function $\Psi(\mathbf{F})$, where $\mathbf{F} = \frac{\partial \mathbf{x}}{\partial \mathbf{X}}$ is the deformation gradient. The mapping $\mathbf{x}(\mathbf{X}, t)$ defines the current configuration at time $t$ relative to the reference configuration $\mathbf{X}$. To ensure material objectivity and to naturally incorporate material symmetry, we use the right Cauchy-Green strain tensor $\mathbf{C} = \mathbf{F}^{T} \mathbf{F}$ as the primary strain measure for finite strain modeling.

In this following, we model the PolyJet 3D printed materials as isotropic and nearly incompressible based on their uniform nature and measured Poissons's ratios close to 0.5. Incompressibility, given by $J = \det(\mathbf{F}) = \det{(\mathbf{C})}^{1/2} = 1$, can be enforced exactly using a augmented energy term $\Psi^{\text{constraint}} = p(J - 1)$, where $p$ is the pressure Lagrange multiplier. Alternatively, it can be approximated using a penalty method with a volumetric term $\Psi^{\text{vol}} = \frac{1}{2}\kappa(J - 1)^2$, where $\kappa$ is a large bulk modulus. We focus on modeling the isochoric response of the material. The isochoric part of the deformation is described using the modified right Cauchy-Green tensor $\bar{\mathbf{C}} = J^{-2/3} \mathbf{C}$, which removes the volumetric component. The first and second isochoric invariants are defined as $\bar{I}_1 = \text{tr}(\bar{\mathbf{C}})$ and $\bar{I}_2 = \frac{1}{2} \left[ (\text{tr}(\bar{\mathbf{C}}))^2 - \text{tr}(\bar{\mathbf{C}}^2) \right]$. When $J = 1$, we see that $\bar{I}_{1} = I_1$ and $\bar{I}_{2} = I_2$.

We express the strain energy as a function of the isochoric invariants \( \bar{I}_1 \), \( \bar{I}_2 \), and parameter $\bm{c}$ which captures variations in material composition. To ensure convexity with respect to the strain invariants while allowing non-convex dependence on ${c}$, we use a pICNN. The network predicts a scalar-valued strain energy \( \hat{\Psi}_{\text{NN}}(\bar{I}_1, \bar{I}_2, {c}) \). To satisfy the normalization condition, which requires that both the strain energy and stress are zero when \( \mathbf{C} = \bm{1} \), we follow the approach proposed by Linden et. al.,~\cite{linden2023neural} to obtain the final strain energy prediction.
\begin{equation}
\hat{\Psi}(\bar{I}_1,\bar{I}_2,J,{c}) = \hat{\Psi}^{NN}(\bar{I}_1,\bar{I}_2,{c}) - \hat{\Psi}^{NN}(3,3,{c}) + \hat{\Psi}^{vol}(J)\\
\end{equation}
or,
\begin{equation}
\hat{\Psi}(\bar{I}_1,\bar{I}_2,J,{c}) = \hat{\Psi}^{NN}(\bar{I}_1,\bar{I}_2,{c}) - \hat{\Psi}^{NN}(3,3,{c}) - \hat{\Psi}^{constraint}(J).
\end{equation}
It can be shown that the stress normalization method proposed by Linden et. al.,~\cite{linden2023neural} is automatically satisfied when the neural network free energy function $\hat{\Psi}_{\text{NN}}$ is formulated using the isochoric strain invariants $\bar{I}_1$ and $\bar{I}_2$. 

The part of the 2nd of the 2nd Piola-Kirchoff (2nd-PK) stress without the incompressibility constraint can then be calculated by,
\begin{equation}
\mathbf{S}' = 2\frac{\partial \Psi}{\partial \mathbf{C}} = 2\Bigg[\frac{\partial \Psi}{\partial \bar{I}_{1}} \frac{\partial \bar{I}_{1}}{\partial \mathbf{C}} + \frac{\partial \Psi}{\partial \bar{I}_{2}}\frac{\partial \bar{I}_{2}}{\partial \mathbf{C}}\Bigg],
\end{equation}
where 
\begin{equation}
\frac{\partial \bar{I}_{1}}{\partial \mathbf{C}} = J^{-2/3}\Big[\bm{1} - \frac{1}{3}\text{tr}(\mathbf{C})\mathbf{C}^{-1} \Big],
\end{equation}
and 
\begin{equation}
\frac{\partial \bar{I}_{2}}{\partial \mathbf{C}} = \bar{I}_{1}\frac{\partial \bar{I}_{1}}{\partial \mathbf{C}} - J^{-4/3}\mathbf{C} + \frac{1}{3}J^{-4/3}\text{tr}(\mathbf{C}^{2})\mathbf{C}^{-1}.
\end{equation}
The Cauchy stress without the volumetric constraint can be obtained from the 2nd-PK stress by  $\bm{\sigma'} = J^{-1}\mathbf{F}\mathbf{S'}\mathbf{F}^{T}$.

\subsubsection{Partially Input Convex Neural Network}
\label{sec:model_picnn}

\begin{figure}[ht]
\centering
\includegraphics[width=\textwidth]{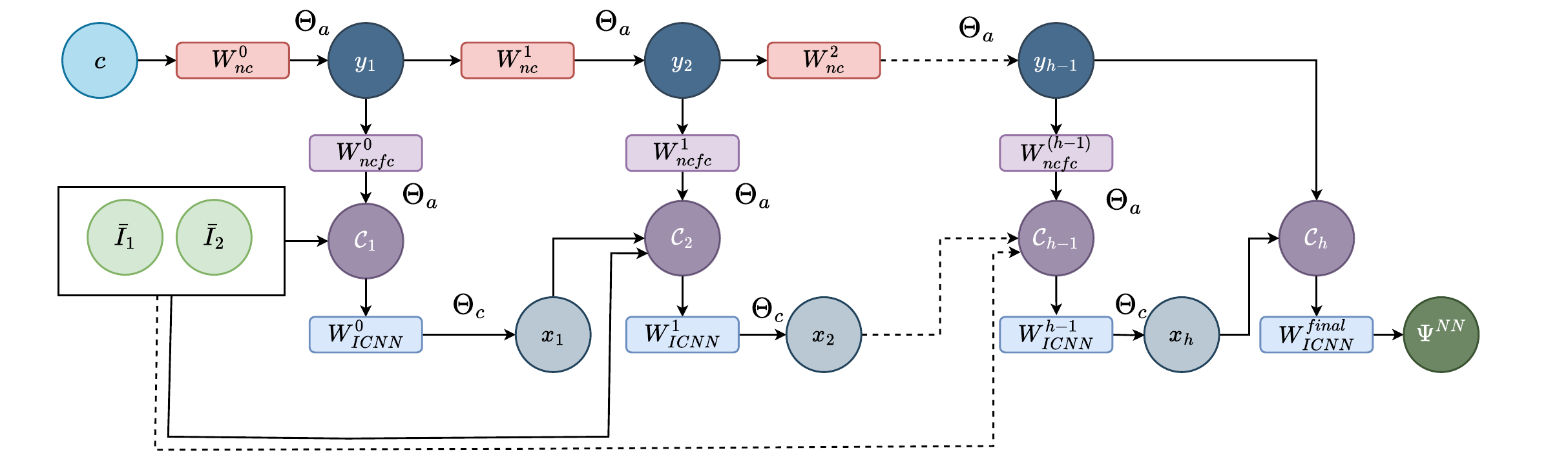}
\caption{Architecture of the proposed partial input convex neural network (pICNN), where strain invariants serve as inputs to the convex part, and material composition features are provided to the non-convex part.}
\label{fig:picnn}
\end{figure}

Let the input invariants be denoted by $\mathbf{\bar{I}}$, and let the material composition be represented by ${c}$. We construct a partially input convex neural network (pICNN) architecture that enforces convexity of the output with respect to the invariants $\mathbf{\bar{I}}$ while allowing for general (non-convex) dependence on the mechanical properties ${c}$. The proposed model consists of two neural networks: a non-convex path that processes ${c}$, and a convex path that processes $\mathbf{\bar{I}}$. These two networks are coupled through intermediate feature fusion operations at each hidden layer, allowing the network to retain convexity with respect to strain measures while capturing material-specific nonlinearities governed by the mechanical properties (see Figure \ref{fig:picnn}). 

Our formulation of the pICNN architecture introduces two key modifications compared to previous approaches used for modeling hyperelastic materials ~\cite{jadoon2024inverse}. These changes are designed to improve computational efficiency and enhance training stability. First, we eliminate all bias terms from the network, which reduces the number of trainable parameters and promotes faster convergence. Second, we use passthrough connections to reintroduce the input strain invariants in all subsequent layers, but do so without any trainable weights. This simplifies the network structure, minimizes redundancy, and ensures that convexity with respect to the strain invariants is maintained.

\par
In the non-convex path, the material parameters ${c}$ are passed through a series of fully connected layers that are implemented via $L_{0}$-regularized linear maps~\cite{louizos2017learning}, each followed by a \textit{softplus} activation function $\Theta_a$. Specifically, the non-convex representations are updated recursively according to
\begin{equation}
  \mathbf{y}_1=\Theta_a\left(\mathbf{W}_{n c}^{(0)}{c}\right), \quad \mathbf{y}_h=\Theta_a\left(\mathbf{W}_{n c}^{(h-1)} \mathbf{y}_{h-1}\right), \quad h=2, \ldots, H  ,
\end{equation}
where $\mathbf{W}_{nc}^{(h)}$ are trainable weights that map the material parameters ${c}$ to the ouput. The intermediate vectors $\mathbf{y}_h$ are then non-linearly transformed into output features $\mathbf{\mathcal{C}}_h=\Theta_a(\mathbf{W}_{ncfc}^{h} \mathbf{y}_h)$, which are integrated into the convex network.

The convex network computes a hierarchical transformation of the invariants $\mathbf{\bar{I}}$, beginning with an initial output from the non-convex with $\mathbf{c}_1$. The first hidden layer is defined as
\begin{equation}
   \mathbf{x}_1=\Theta_c\left(\mathbf{W}_{ICNN}^{(0)} \mathbf{\bar{I}}+\mathbf{W}_{ncfc}^{(0)} {y}_1\right) 
\end{equation}
and for each subsequent layer $h=1, \ldots, H$, the network computes
\begin{equation}
\mathbf{x}_h=\Theta_c\left(\mathbf{W}_{ICNN}^{(h-1)} \mathbf{\bar{I}}+\mathbf{x}_{(h-1)}+\mathbf{W}_{ncfc}^{(h-1)} y_{(h-1)}\right)
\end{equation}

Here, $\Theta_c$ is a convex and monotonically non-decreasing activation function, such as the \textit{softplus} function. The weights $\mathbf{W}_{ICNN}^{0}$, which connect the invariants directly to the first hidden layer, and the hidden-to-hidden weights $\mathbf{W}_{ncfc}^{(0)}$ are unconstrained, i.e, it can either be positive or negative~\cite{amos2017input}. $\mathbf{W}_{ICNN}^{h}$ is the weight that maps the input to the various hidden layers of the convex network. To ensure convexity of the network output with respect to I, all subsequent input weights for $h \geq 2$ are non-negative.
\par
Finally, the final scalar-valued output, representing the strain energy potential $\Psi$, is computed by a convex linear transformation of the final hidden states:
\begin{equation}
\hat{\Psi}^{NN}=\mathbf{W}_{\text{ICNN}}^{h} (\mathbf{x}_H+{C}_H+\mathbf{\bar{I}}),
\end{equation}
where $\mathbf{W}_{\text{ICNN}}^{h} $ are constrained to be non-negative. No bias terms are included in the formulation. All dense layers are implemented using a $L_0$-sparsification mechanism to promote sparsity and interpretability~\cite{louizos2017learning}. This architecture guarantees the convexity of the strain energy potential $\Psi^{NN}$ with respect to the invariants $\mathbf{\bar{I}}$ while enabling expressive and nonlinear modeling of material behavior through its dependence on the mechanical parameters ${c}$. \\

\subsubsection{Smoothed $L_0$ Sparsification}

$L_0$ sparsification techniques are widely employed to enhance model interpretability and reduce complexity by regularizing model parameters. However, the inherent non-differentiability of the $L_0$ norm poses significant optimization challenges~\cite{fuhg2024extreme}. To address this, we adopt the smoothed $L_0$ sparsification approach proposed by Louizos et al.~\cite{louizos2017learning}. This technique introduces a differentiable, continuous gating mechanism, where each model parameter is controlled by a gate value $z \in [0,1]$, becoming effectively inactive when its corresponding gate value approaches zero, and remaining active when the gate value approaches one. The gating mechanism is implemented via a smooth approximation, ensuring differentiability throughout the optimization process:
\begin{equation}
    \bm{\theta}=\overline{\bm{\theta}} \odot \mathsf{z},
\end{equation}
where $\mathsf{z}=\min (\mathbf{1}, \max (\mathbf{0}, \bar{s}))$, $\odot$ denotes the Hadamard product and $\bm{\theta} = [\bm{W}_{ncfc},\bm{W}_{fc},\bm{W}_{nc}]$. The intermediate gating vector $\overline{\bm{s}}$ is formulated as:

\begin{equation}
\begin{aligned}
& \overline{\bm{s}}=\bm{s}(\zeta-\gamma)+\gamma \mathbf{1} \\
& \bm{s}=\operatorname{sig}((\log \bm{u}-\log (1-\bm{u})+\log \alpha) / \beta)
\end{aligned}
\end{equation}
The hyperparameters $\gamma$, $\beta$, $\zeta$, and $\log \bm{\alpha}$ regulate the smoothness and behavior of the gating function, providing the flexibility required for effective sparsification. Here, $\bm{u}$ is a random vector sampled uniformly from the interval $[0,1]$, with the same dimension as $\mathsf{z}$. From \cite{louizos2017learning}, we set hyperparameters as $\gamma = -0.1$, $\zeta = 1.1$, and $\beta = \frac{2}{3}$, while initializing $\log \bm{\alpha}$ from a Gaussian distribution $\mathcal{N}(0, \sigma)$ with standard deviation $\sigma = 0.01$. Given the stochastic nature of this gating mechanism, we construct a Monte Carlo approximation of the loss function as follows:
\begin{equation}
\begin{aligned}
\mathcal{R}(\overline{\bm{\theta}}) & =\frac{1}{M} \sum_{j=1}^M\left(\frac{1}{N}\left(\sum_{i=1}^N \mathcal{L}\left(\mathrm{PANN}\left(\mathrm{x}_i, \overline{\bm{\theta}} \odot \mathrm{z}^m\right), \mathrm{y}_i\right)\right)\right] \\
& +\lambda \sum_{j=1}^{\bm{\theta}} \operatorname{sig}\left(\log \alpha_j-\beta \log \frac{-\gamma}{\zeta}\right),
\end{aligned}
\end{equation}
where $M$ denotes the number of Monte Carlo samples used, $N$ is the number of data points, $x$ is the input, $y$ is the output to the PANN, and $\lambda$ is a regularization coefficient that controls the trade-off between sparsity and model fit. 
\\
At the inference time, we obtain the final prediction using optimized parameters by multiplying the trained parameters with the gated vector:  $\bm{\theta}^*=\overline{\bm{\theta}}^* \odot \hat{\mathsf{z}}$ , where $\hat{\mathsf{z}}$ is given as follows:
\begin{equation}
\hat{\mathsf{z}}=\min (\mathbf{1}, \max (0, \operatorname{sig}(\log \alpha)(\zeta-\gamma)+\gamma 1)).
\end{equation}
Therefore, this ensures consistent parameter selection, effectively setting inactive parameters close to zero and thus enhancing model interpretability and sparsity at test time.
\subsection{Viscoelastic Model}
\label{sec:model_qlv}
We model the mechanical response of the digital materials (DMs) using a formulation based on the Quasi-Linear Viscoelastic (QLV) approach \citep{wine09MMS,depa&etal14PRSA}. The Cauchy stress takes the general form
\begin{equation}
\bm{\sigma}(t) = \bm{\sigma^{e}}(t) + \int_{0}^{t}D'(t-s)\bm{\sigma^{e}}(t)\,ds
\label{eq:second_Piola_Kirchhoff},
\end{equation}
where $\bm{\sigma}(t)$ is the Cauchy stress at time $t$, $\bm{\sigma^{e}}$ is the instantaneous (hyperelastic) stress response, and $D'(t-s)$ is an integration kernel that accounts for loading history. We define the kernel as:
\begin{equation}
D'(t-s) = -\frac{\gamma}{\tau}e^{-(t-s)/\tau},
\label{Eqn:D_evolution}
\end{equation}
where $\tau$ is the relaxation time and $\gamma$ is the relaxation coefficient. In this study we fix $\tau = 10$ seconds, which is on the scale of observation, while treating $\gamma$ as a composition-dependent material property.Since $\gamma$ and $\tau$ appear together as the ratio $\gamma/\tau$ in equation \ref{Eqn:D_evolution}, the parameters are coupled. Given the limited amount of training data fixing $\tau$ helps reduce the complexity of optimizing the viscoelastic model. To capture the relationship between composition and viscoelastic behavior, we design a MLP to model $\gamma$ as a function of material composition vector $\bm{c}$. The MLP applies softplus activations in hidden layers and a final sigmoid to constrain $\gamma \in [0,1]$. Constraining $\gamma$ ensures that optimization results in values that are usable. If the network had $L$ layers with no bias the network is defined as
\begin{equation}
\hat{\gamma}= \Theta_{\operatorname{sig}}\left(\bm{W}_L \cdot \Theta_{\operatorname{softplus}}\left(\bm{W}_{L-1} \cdot \Theta_{\operatorname{softplus}}\left(\ldots \Theta_{\operatorname{softplus}}\left(\bm{W}_1 \cdot \bm{c}\right) \ldots\right)\right)\right),
\end{equation}
where the $\Theta_{\operatorname{sig}}$ is the sigmoid activation function, $\Theta_{\operatorname{softplus}}$ is the softplus activation function, and the model parameters $\bm{W}$.

\subsection{Modeling Uniaxial Tension}
\label{sec:model_tension}
Assuming material incompressibility, the deformation gradient for uniaxial tension is written as
\begin{equation}
\mathbf{F}_{U}=\left(\begin{array}{ccc}
\lambda(t) & 0 & 0\\
0 & \lambda(t)^{-\frac{1}{2}} & 0\\
0 & 0 & \lambda(t)^{-\frac{1}{2}}
\end{array}\right),
\end{equation}
and the right Cauchy-Green strain tensor can be written as
\begin{equation}
\mathbf{C}_{U}=\left(\begin{array}{ccc}
\lambda(t)^{2} & 0 & 0\\
0 & \lambda(t)^{-1} & 0\\
0 & 0 & \lambda(t)^{-1}
\end{array}\right),
\end{equation}
where $\lambda(t)$ is the axial stretch at time $t$. With the incompressible assumption, only the axial stretch measurement is needed to fully define the deformation gradient. This gives, $I_{1} = \bar{I}_{1} = \lambda^{2} + 2\lambda^{-1}$, $I_{2} = \bar{I}_{2} = 2\lambda + \lambda^{-2}$, and $J=1$. 

For uniaxial tension, only the axial stress $\sigma_{11}$ is expected to be nonzero with $\sigma_{22} = \sigma_{33} = 0$. The full Cauchy stress is given by $\mathbf{\sigma} = \mathbf{\sigma'} + p\bm{1}$ where $\mathbf{\sigma'}$ is the Cauchy stress without the volumetric constraint and $p$ is pressure. For uniaxial tension, the deviatoric stress has one distinct component, $\sigma_{11}'$, equal transverse components, $\sigma_{22}' = \sigma_{33}'$ and all shear components equal to zero. Using these conditions, the pressure is determined as $p = -\sigma_{22}'=-\sigma_{33}'$, which allows the full stress tensor to be recovered.

\subsection{Modeling Torsion}
\label{sec:model_torsion}
To model the deformation from the torsion test, consider a rod of length $L$, and outer radius $R$ that is axially constrained and twisted about its axis by $\varphi$ radians. We define a polar coordinate system $\{\hat{\bm{r}},\hat{\bm{\theta}},\hat{\bm{z}}\}$ and the corresponding deformation gradient at a given radial position $r$ is:
\begin{equation}
\mathbf{F}_{T}=\left(\begin{array}{ccc}
1 & 0 & 0\\
0 & 1 & \frac{\varphi}{L}\,r\\
0 & 0 & 1
\end{array}\right), 
\label{eq:defgrad_twist}
\end{equation}
and the right Cauchy-Green strain tensor is 
\begin{equation}
\mathbf{C}_{T}=\left(\begin{array}{ccc}
1 & 0 & 0\\
0 & 1 & \frac{\varphi}{L}\,r\\
0 & \frac{\varphi}{L}\,r & \big(\frac{\varphi}{L}\,r\big)^{2} + 1
\end{array}\right).
\end{equation}
In this case, the first and second strain invariants are equal with $I_{1}=\bar{I}_{1}=I_{2}=\bar{I}_{2}=3+\big(\frac{r}{L}\,\varphi\big)^{2}$, and $J = \sqrt{I_{3}} = 1$. If the strain energy density is a function of invariants $I_{1}$ and $I_{2}$, it can be shown that the Cauchy stress component in the $z\theta$ direction at a given radial position $r$ is given by
\begin{equation}
\sigma_{z\theta}(r) = 2\frac{\varphi}{L}\,r\Bigg[\frac{\partial \Psi}{\partial I_{1}} + \frac{\partial \Psi}{\partial I_{2}}\Bigg].
\end{equation}
We follow Horgan and Saccomandi~\cite{horgan1999simple} to obtain the axial torque, which is the main measured quantity in our torsion experiments:
\begin{equation}
T = 2\pi\int_{0}^{R} \sigma_{z\theta}(r) r^{2} \,dr, 
\end{equation}
where $T$ is the torque and $R$ is the outer radius of the rod.

\section{Results and discussion }\label{sec:Results-and-discussion}
\subsection{Experimental Results}
To begin characterizing the digital materials (DM) used in this study, we first quantified the proportions of the base polymers supplied for multi-material printing. The base polymers, Agilus, RGD515, and RGD531 are proprietary formulations provided by Stratasys and serve as the building-blocks for DM compositions through voxel- and drop-wise-level blending. Agilus is designed to exhibit an elastomeric response, characteristic of soft, rubber-like materials, while the combination of RGD515 and RGD531 referred to as Digital ABS demonstrates a more glassy, rigid behavior. Although each base polymer may consist of multiple undisclosed components, they represent the primary feedstocks used in this PolyJet printing process. To determine their relative amounts in each DM composition, we printed 10x10x10 $\text{cm}^{3}$ specimens and recorded the mass of each base polymer consumed. Table~\ref{tbl:percent_agilus_digital_abs} summarizes the mass percentage of base polymers for each DM composition.

\begin{table}[ht]
\small
\caption{Mass percentages of Agilus, RGD515, and RGD531 for various digital material (DM) compositions. The combination of RGD515 and RGD531 is referred to as Digital ABS.}
\label{tbl:percent_agilus_digital_abs}
\begin{tabular*}{\textwidth}{@{\extracolsep{\fill}}llll}
\hline
      &           & \multicolumn{2}{c}{Digital ABS} \\
      & \% Agilus & \% RGD515      & \% RGD531      \\ \hline
A     & 75.2      & 11.8           & 13.0           \\
DM-40 & 74.8      & 11.7           & 13.5           \\
DM-50 & 74.1      & 11.7           & 14.2           \\
DM-60 & 72.8      & 11.7           & 15.4           \\
DM-70 & 70.3      & 11.7           & 18.0           \\ \hline
\end{tabular*}
\end{table}

Notably, across the different DM compositions, the amount of RGD515 remains constant while the proportion of RGD531 and Agilus vary. Although both RGD515 and RGD531 comprise the Digital ABS component, only RGD531 is adjusted to tune the mechanical properties of the final material. Across the five DM compositions, A has the highest Agilus content and lowest Digital ABS, while DM-70 has the opposite. Despite only a 5\% shift in composition between A and DM-70, each composition step has an approximate 10-unit increase in Shore A hardness, highlighting how small formulation changes are used to produce significant mechanical differences.

\begin{figure}[ht]
\begin{centering}
\includegraphics[width=\textwidth]{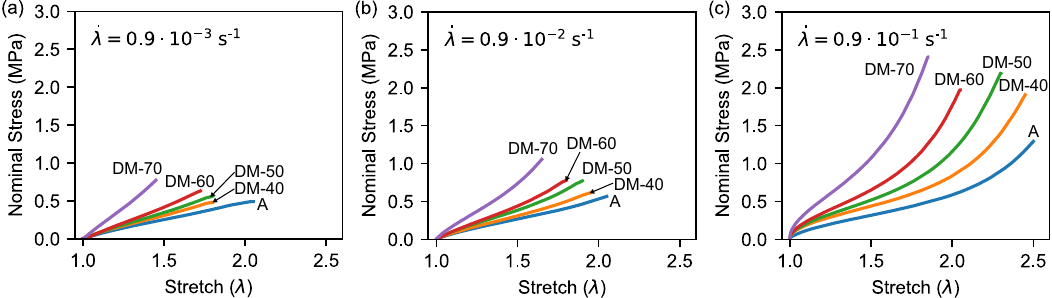}
\end{centering}
\caption{Experimental uniaxial tension data showing nominal stress versus stretch for various digital material (DM) compositions tested at a stretch rate ($\dot{\lambda}$) of (a) $0.9 \cdot 10^{-3}$, (b) $0.9 \cdot 10^{-2}$, and (c) $0.9 \cdot 10^{-1}$ $\text{s}^{-1}$.}
\label{fig:tension_rates}
\end{figure}

To more precisely quantify the mechanical properties of the DMs, we performed both uniaxial tension and torsion tests across multiple compositions and deformation rates. While the Shore A hardness values provide a basic indication of stiffness, they do not fully capture the rate-dependent and nonlinear behavior observed in these materials. Figure \ref{fig:tension_rates} shows the nominal stress versus stretch ($\lambda$) from uniaxial tension tests conducted across two decades of strain rate.

All five DM compositions demonstrate qualitatively similar stress-stretch behavior, with clear differences in stiffness, strength and extensibility. Materials with higher Digital ABS content exhibit greater small-strain stiffness (Young's modulus), higher stress at break, and reduced elongation before failure. For each applied stretch rate, each successive composition from Agilus to DM-70 exhibit higher stress at the same stretch level, with DM-70 showing the highest stresses and Agilus the lowest. 

At the highest applied stretch rate ($\dot{\lambda} = 0.9 \cdot 10^{-1}$ $\text{s}^{-1}$), the curves display a characteristic upturn in stress at large deformations. This behavior is typical of polymeric materials and is generally attributed to the deformation-induced alignment and limited extensibility of polymer chains as they approach their stretch limit. Additionally, as the Digital ABS content increases, particularly evident in DM-70, the curves develop a more pronounced initial bump, characterized by an initially steep response followed by a slight softening before the final upturn. This feature may indicate a shift from a purely elastomeric response to one influenced by glassy polymer behavior, potentially reflecting localized yielding or microstructural rearrangement associated with the stiffer Digital ABS component.

\begin{figure}[ht]
\begin{centering}
\includegraphics[width=\textwidth]{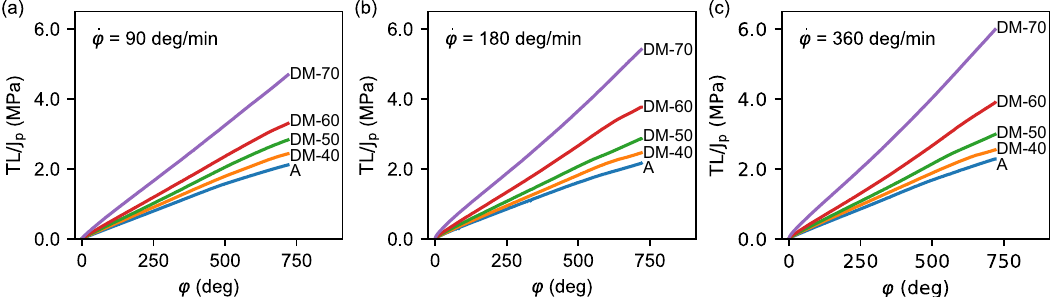}
\end{centering}
\caption{Experimental torsion data showing $TL/J_{p}$ (MPa) versus rotation angle $\varphi$ (deg) for various digital material (DM) compositions tested at twist rates of (a) $\dot{\varphi} = 90$ deg/min, (b) $\dot{\varphi} = 180$ deg/min, and (c) $\dot{\varphi} = 360$ deg/min.}
\label{fig:torsion_rates}
\end{figure}

In order to show the mechanical behavior under torsion, the torque is multiplied by the length of the specimen, $L$, and normalized by the polar moment of inertia $J_{p} = (\pi/2)R^{4}$ where $R$ is the radius of the specimen. In Figure \ref{fig:torsion_rates}, we plot $TL/J_{p}$ as a function of twist angle $\varphi$ for $\varphi\leq720^{\circ}$. Using this representation, the slope under small angles gives the shear modulus $\shear$. For all digital materials and twist rates, the $TL/J_{p}$ vs. $\varphi$ response is approximately linear, despite the nonlinear stress-strain behavior observed in tension. As the Digital ABS content increases, so does the shear modulus, with a clear trend from Agilus to DM-70.

\begin{figure}[ht]
\begin{centering}
\includegraphics[width=\textwidth]{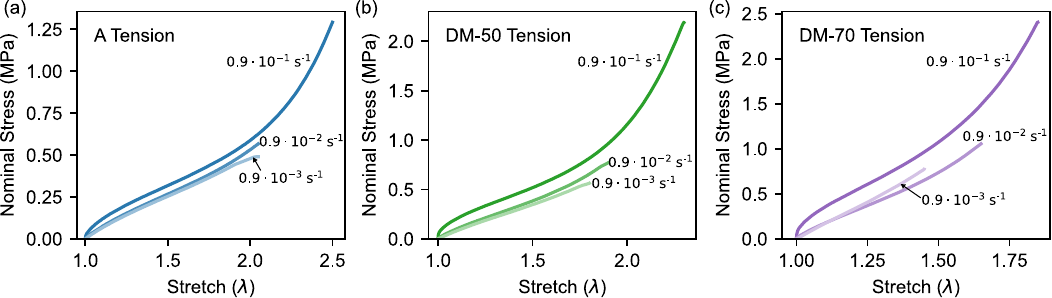}
\end{centering}
\caption{Select uniaxial tension plots for (a) A, (b) DM-50, and (c) DM-70 showing nominal stress versus stretch. Different stretch rates ($\dot{\lambda} = 0.9 \cdot 10^{-3}$, $0.9 \cdot 10^{-2}$, and $0.9 \cdot 10^{-1}$ $\text{s}^{-1}$) are shown together for each composition to highlight effects of deformation rate.}
\label{fig:tension_3dms}
\end{figure}

The effect of material composition on rate-dependent behavior becomes more evident when the stress-stretch responses at all three stretch rates are plotted together for each material. Figures \ref{fig:tension_3dms}(a-c) illustrates this comparison for three selected compositions: A, DM-50, and DM-70. Interestingly, the A composition shows almost no viscoelastic effect with no significant difference between the three rates (Figure \ref{fig:tension_3dms}(a). In contrast, for DM-50 and DM-70 under tension, the two slower stretch rates ($\dot{\lambda} = 0.9 \cdot 10^{-2}$ and $\dot{\lambda} = 0.9 \cdot 10^{-3}$ $\text{s}^{-1}$) result in lower stresses compared to the highest rate. Notably, the stress responses at these two slower rates are nearly identical, suggesting that both fall within the quasi-static regime and that equilibrium is effectively reached at rates below $\dot{\lambda} = 0.9 \cdot 10^{-2}$ $\text{s}^{-1}$. 

\begin{figure}[ht]
\begin{centering}
\includegraphics[width=\textwidth]{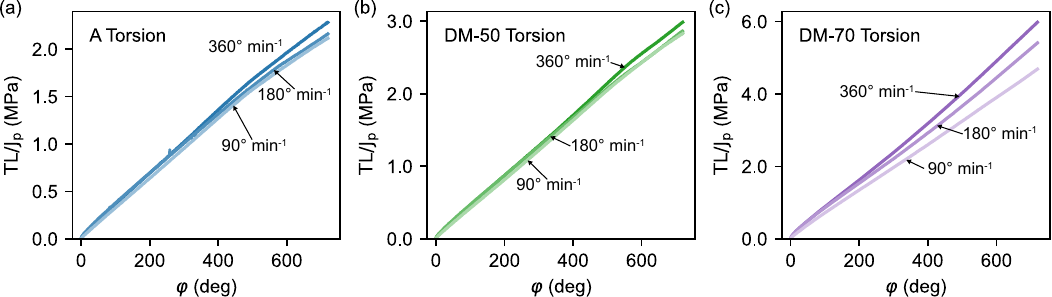}
\end{centering}
\caption{Select torsion tests for (a) A, (b) DM-50, and (c) DM-70 showing $TL/J_{p}$ (MPa) versus rotation angle $\varphi$ (deg). Different twist rates ($\dot{\varphi}$ = 360, 180, 90 deg/min) are shown together for each composition to highlight effects of deformation rate.}
\label{fig:torsion_3dms}
\end{figure}

For torsion, we compare the effect of twist rate across three selected compositions: A, DM-50, and DM-70 as shown in Figures \ref{fig:torsion_3dms}(a-c). Similar to the tension results, Agilus exhibits almost no rate dependence in torsion across the tested twist rates, confirming its largely rate-independent response. For DM-50, the torque-twist curves show minimal difference from 90 deg/min to 360 deg/min, while DM-70 displays some increase in torque at the higher twist rate. 

Although torsion appears less sensitive to deformation rate based on the minimal differences observed between twist rates, this is explained by the deformation rates imposed during testing. The engineering shear strain rate in torsion, given by $(r/L)\dot{\varphi}$, reaches approximately 0.0098 $\text{s}^{-1}$ at the outer radius under the fastest twist rate. This value is comparable to the middle strain rate used in the tension tests (0.009 $\text{s}^{-1}$). Since the tension results showed that rates at or below this level fall within the quasi-static regime, it follows that the torsion tests were also conducted under quasi-static conditions, accounting for the limited rate sensitivity observed in the torsion experiments.

Lastly, the Poisson's ratios of the different materials were measured by \citet{levin&cohen23AMT} to be $\nu\geq 0.47$. 
Since the ratio between the bulk modulus $\bulkm$ and the shear modulus $\frac{\bulkm}{\shear}=\frac{2\left(1+\nu\right)}{3\left(1-2\nu\right)} \geq 16.3$, it is reasonable to treat these materials as incompressible.

\subsection{Numerical Implementation and Model Results}

\subsubsection{Data Preparation}

To model the mechanical response of the digital materials (DMs), we combine a physics-augmented neural network for the instantaneous hyperelastic response with a quasi-linear viscoelastic model to capture rate-dependent behavior. The partial input convex neural network (described in Section \ref{sec:model_picnn}) enforces convexity of the strain energy function with respect to the strain invariant inputs $\bar{I}_1$ and $\bar{I}_2$, which are computed directly from the experimental deformation data under the assumption of incompressibility (Sections \ref{sec:model_tension} and \ref{sec:model_torsion}). We use the non-convex part of the network to adapt the model to varying material compositions. Here, the proportion of Agilus versus Digital ABS of the DMs is provided as the non-convex input $c$, as detailed in Table \ref{tbl:c_input}. 

\begin{table}[ht]
\caption{Digital material composition input ($c$) values for the non-convex part of the pICNN.}
\label{tbl:c_input}
\begin{tabular*}{\textwidth}{@{\extracolsep{\fill}}lll}
\hline
Material & \% Digital ABS & $c$      \\ \hline
A     & 24.8           & 0      \\
DM-40 & 25.2           & 0.1755 \\
DM-50 & 25.9           & 0.4669 \\
DM-60 & 27.2           & 1.0    \\
DM-70 & 29.7           & 2.0895 \\ \hline
\end{tabular*}
\end{table}

The composition parameter ($c$) is scaled to ensure the magnitude is consistent with the other model inputs. In our case, the magnitude of the strain invariants is between 1 and ~7. Specifically, we made $c=0$ correspond to the Digital ABS content of A, and made $c=1$ correspond to DM-60 with values extending beyond 1 for DM-70. 

For model training, we used an 80/20 train-validation split on both the uniaxial tension and torsion experiments. One intermediate composition (DM-40) was held out entirely for testing of interpolation performance, and one higher composition (DM-70) was held out to test extrapolation. During training, we used a total of 4 batches per experiment, with a batch size of 8 for the uniaxial tension data and 4 for the torsion data.

\subsubsection{PANN architecture and hyperparameter tuning}

In this subsection, we empirically perform the hyperparameter search for the pICNN architecture and other hyperparameters such as learning rate, weight decay, number of epochs and number of training data. Here, the model architecture for the pICNN strain energy function consists of two fully connected layers with $30$ nodes each for the convex part of the network, and two layers with $5$ nodes each for the non-convex portion. This serves as the starting point for training, with a total of $1,357$ weights in the convex part and $75$ weights in the non-convex part, which includes the connections between the non-convex and convex layers. After sparsification via  $L_{0}$-regularization, the trained model consists of $17$ parameters of convex weights and $16$ for the non-convex part, which includes the connections. For the MLP network that maps the composition input $c$ to the parameter $\gamma$ used in the quasi-linear viscoelastic model, we use a network with one hidden layer of $8$ nodes given the low dimensionality of this mapping. The models were trained using Adam optimizer~\cite{kingma2014adam} with the following learning rate for the pICNN and MLP: $0.005$ and $0.001$, respectively.

\subsubsection{Training Procedure}

In this study, we train the pICNN model on experimental data from uniaxial tension and torsion tests. We used a multi-step training approach due to the complexity of optimizing models for viscoelasticity and torsion, particularly because of the integration over the loading history. Single-step optimization often struggles with complex parameter landscapes and may become trapped in local minima. The multi-stage method enables model weights to more effectively explore the parameter space, improving convergence and generalization, and addressing inherent nonlinearities and the existence of non-unique solutions resulting from the integration involved in viscoelastic and torsional behaviors. The loss function, defined in Equation~\ref{eqn:loss_multi}, is evaluated over seven sequential steps during training (steps 1-7) as indicated in Algorithm~\ref{algo:train}, where $\bm{\theta}$ denotes the model parameters. Specifically:
\begin{itemize}
    \item 

Data fidelity (Steps 1 to 6 in Algorithm~\ref{algo:train}): For each experiment $k$ out of $N$ total, we compute the mean squared error (MSE) using the squared $L_2$ norm, $|\cdot|_{2}$ ($p=2$ in Equation~\ref{eqn:loss_multi}) without penalty term ($\alpha_{fc}||\theta_{fc}||_{0} = \alpha_{nc}||\theta_{fc}||_{0} = \alpha_{ncfc}||\theta_{ncfc}||_{0} = 0$), separately for tension and torsion measurements. To mitigate scale differences, stress and torque values are normalized by their experiment-specific minima and maxima. Separate weights $\alpha_{tension}$ and $\alpha_{torsion}$ balance the contributions from the two datasets.
\item 
Sparsity regularization (Step 7 in Algorithm~\ref{algo:train}): We apply an $L_0$ penalty (in Equation~\ref{eqn:loss_multi}) to encourage network sparsity. This regularization is imposed independently on three parameter groups of the pICNN fully convex (fc), non-convex (nc), and convex-to-non-convex connecting (ncfc) with respective weights $\alpha_{fc}$, $\alpha_{nc}$, and $\alpha_{ncfc}$.
\end{itemize}
Overall, this formulation ensures accurate fitting across experiments (via steps 1-6) while promoting a sparse parameterization (via step 7).

\begin{equation}
\label{eqn:loss_multi}
\begin{aligned}
\mathcal{L}(\bm{\theta};\bar{I}_{1},\bar{I}_{2},c) = \alpha_{tension}\frac{1}{N_{tension}}\sum_{k=1}^{N_{tension}}||\frac{1}{\max(\sigma_{11}^{(k)})-\min(\sigma_{11}^{(k)})}(\sigma_{11}^{(k)} - \hat{\sigma}_{11}^{(k)})||_{2}^{2} + \\
\alpha_{torsion}\frac{1}{N_{torsion}}\sum_{k=1}^{N_{torsion}}||\frac{1}{\max(T^{(k)}) - \min(T^{(k)})}(T^{(k)} - \hat{T}^{(k)})||_{2}^{2} + \\ \alpha_{fc}||\bm{\theta}_{fc}||_{0} +\alpha_{nc}||\bm{\theta}_{nc}||_{0} + \alpha_{ncfc}||\bm{\theta}_{ncfc}||_{0} 
\end{aligned}
\end{equation}

Training was staged to progressively introduce different model components. The instantaneous stress response (PANN) was first trained on the fastest uniaxial tension rate, followed by training the QLV $\gamma$ parameter (MLP) using the full set of tension experiments across all three rates. The PANN and MLP were then updated through alternating steps. The QLV model was trained only on tension data, as the torsion data lie primarily in the quasi-static regime. Once the rate-dependent response was established, torsion data were gradually introduced to capture multiaxial behavior. Finally, $L_{0}$-regularization was applied to promote sparsity in the PANN. Unless otherwise noted, Steps 1-5 were trained using only the uniaxial tension loss 
(\(\alpha_{\text{tension}} = 1\)) with all other loss weights set to zero 
($\alpha_{\text{torsion}} = \alpha_{\text{fc}} = \alpha_{\text{nc}} = \alpha_{\text{ncfc}} = 0$).

\begingroup
\setstretch{1.0} 
\begin{algorithm}[!ht]
\caption{Training procedure for the pICNN-QLV model}
\label{algo:train}
\begin{algorithmic}[noend,1]
 \State \textbf{Input:}  Uniaxial tension data at $\dot\lambda=0.9\times10^{-1}, 0.9\times10^{-2}, 0.9\times10^{-3}\,\mathrm{s}^{-1}$; torsion at $\dot\varphi=360\,\mathrm{deg/min}$
\State \textbf{Initialize:}  Trained pICNN and QLV models
  \State Initialize pICNN ($[\bm{W}_{ICNN},\bm{W}_{nc},\bm{W}_{ncfc}]$) and QLV parameters ($\bm{W}$)
  \State \textbf{Step 1:} 
    \For{epoch = 1 to 1000}
  \State Train pICNN on fastest stretch rate to fit $\bm\sigma_e'$
  \Comment{Trained on uniaxial tension data}
  \EndFor
  \While{not converged}
    \State \textbf{Step 2\&4:} 
    \For{epoch = 1 to 200}
    \State Freeze pICNN; train QLV across all rates 
    \State Predict $\bm\sigma(t)$
        \Comment{Optimized using uniaxial tension data across all $3$ strain rates}
    \EndFor
    \State \textbf{Step 3\&5:}
    \For{i=1 to 200}
    \State  Freeze QLV; fine-tune pICNN on fastest rate for full time-dependent stress
    \State Predict $\bm\sigma(t)$
    \Comment{Fine-tuned using uniaxial tension data at the fastest stretch rate}
    \EndFor
  \EndWhile
  \For{epoch = 1 to $100$}
  \State \textbf{Step 6:} 
  \State Ramp in torsion: increase $\alpha_{\rm torsion}$ from 0 to 0.1 (keep $\alpha_{\rm tension}=1$, others 0)
  \State Predict $\bm\sigma(t)$
      \Comment{Optimized using torsion data}
  \EndFor
  \For{epoch = 1 to $1000$}
  \State \textbf{Step 7:} Apply $L_0$-regularization: ramp $\alpha_{\rm fc}\to5\times10^{-4}$, $\alpha_{\rm nc},\alpha_{\rm ncfc}\to10^{-6}$
    \State Predict $\bm\sigma(t)$
    \Comment{Obtain sparse regularized expression}
  \EndFor
  \State \textbf{Output:} Trained pICNN and QLV models
\end{algorithmic}
\end{algorithm}
\endgroup

\subsubsection{Evaluation metrics}
To evaluate the performance of the algorithm, we define the set of samples $\{\bm{x}^{i}, {y}^{i}\}_{i=1}^{N}$, where $\bm{x}$ is the input data, ${y}$ is the output data, and $N$ is the number of test data. 
In the context of this work, the output ${y}$ corresponds to the axial Cauchy stress \(\sigma_{11}^i\) for uniaxial tension experiments and the torque \(T^i\) for torsion experiments.
We consider the following evaluation metrics for our trained models for different datasets.
\begin{itemize}
\item 
Coefficient of determination ($R^2$ score):
\begin{equation}
R^2 = 1 - \frac{\sum_{i=1}^{N} (\hat{y}^i - y^i)^2}{\sum_{i=1}^{N}(y^i - \bar{y})^2},
\end{equation}
where $\hat{y}$ is the model prediction and $\bar{y}$ is the mean of the experimental data.

\item Symmetric mean absolute percentage error (sMAPE):
\begin{equation}
\text{sMAPE} = \frac{100}{N}\sum_{i=1}^{N} \frac{|\hat{y}^i - y^i|}{|\hat{y}^i| + |y^i|}.
\end{equation}
We use a modified version of sMAPE, where the error is bounded between 0\% and 100\%. This metric is preferred over the standard mean absolute percentage error (MAPE) because it handles values near zero more reliably, which is important since the stress-stretch and torque-rotation curves start at zero. To avoid division by zero, we filter out data points where the experimental value is exactly zero.
\end{itemize}

\subsubsection{Modeling Performance and Predictions}

\begin{figure}[ht]
\begin{centering}
\includegraphics[width=\textwidth]{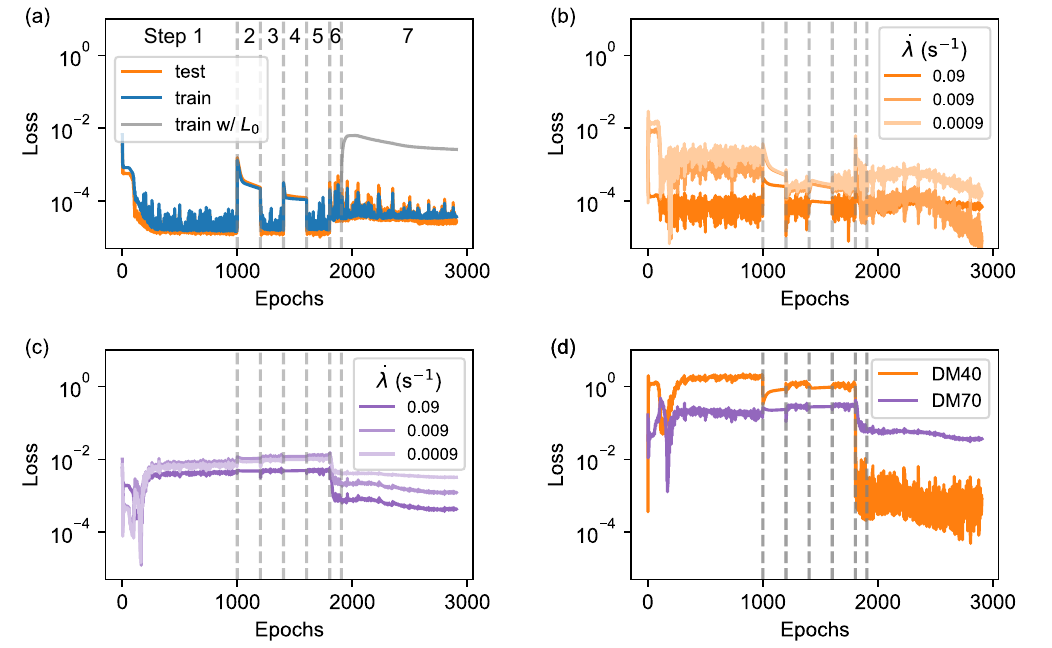}
\end{centering}
\caption{Training and testing loss across model training steps 1-7. (a) The blue line shows the combined loss for the training experiments, and the orange line represents a 20\% subset held out to assess the model's interpolation within the training experiment. The gray line includes the $L_{0}$-regularization in the training loss. (b) Held-out test loss for DM-40 tension experiments, evaluating interpolation between material composition values. (c) Held-out test loss for DM-70 tension experiments, evaluating composition extrapolation. (d) Held-out test loss for DM-40 and DM-70 torsion experiments at 360\textdegree/min.}
\label{fig:loss}
\end{figure}

Throughout the staged training process, both training and test losses progressively improved as new model components and datasets were introduced. Figure \ref{fig:loss}(a) shows the combined training loss, the test loss on a 20\% subset held out from the training experiments, and the training loss with the $L_0$-regularization term included. The largest reduction in training loss occurs during Step 1, when the pICNN model is trained on the fastest-rate tension data ($\dot{\lambda} = 0.09\ \text{s}^{-1}$) to fit the instantaneous stress $\bm{\sigma}_{e}'$, capturing the baseline elastic behavior. In Step 2, an initial increase in training and test loss is observed with the addition of slower-rate tension data ($\dot{\lambda} = 0.009\ \text{s}^{-1}$ and $0.0009\ \text{s}^{-1}$). In this step the full viscoelastic stress $\bm{\sigma}'$ is fit. Updating the MLP that maps material composition $c$ to the QLV $\gamma$ parameter reduces the loss across all loading rates.

During the alternating updates between the pICNN and QLV components (Steps 3-5), loss associated with slower rates continued to decrease, while the fastest-rate loss remained stable. The introduction of torsion data in Step 6 caused a slight increase in total loss, which was progressively reduced through Steps 6 and 7. Step 7 represents the final training stage, incorporating all tension rates and torsion data at $360$ deg/min. Although $L_0$ regularization was gradually introduced by increasing the $\alpha$ parameters over 1000 epochs, the regularization loss is initially large, leading to a noticeable jump followed by a steady decline as the number of pICNN parameters is reduced. Importantly, the loss excluding the $L_0$ term remained stable throughout, indicating that model sparsification was achieved without compromising the fit to the training data. The test loss on the 20\% subset held out from the training experiments (orange line Figure \ref{fig:loss}(a) closely tracked the training loss throughout all stages of the training process, indicating that the model achieved good interpolation performance within the experimental data used for training.

Figures \ref{fig:loss}(b-d) present the test loss for material compositions fully excluded from training. Figure \ref{fig:loss}(b) shows DM-40 tension experiments, evaluating interpolation across composition values $c$, while Figure \ref{fig:loss}(c) presents DM-70 tension experiments, assessing extrapolation beyond the training range. Figure \ref{fig:loss}d displays torsion experiments for DM-40 and DM-70. The DM-40 test loss closely followed the training loss, with the largest reduction in step 1 and continued improvements in steps 2-5 as slower loading rates and viscoelastic behavior was incorporated. Upon introducing torsion data in step 6, the torsion loss for DM-40 also decreased sharply, indicating effective multi-axial learning. Application of $L_0$-regularization in step 7 reduced model complexity without increasing test loss, confirming the model's ability to interpolate within the material composition range present in the training data. In contrast, the DM-70 test loss remained high through steps 1-6, with only minor improvement during step 7. The final DM-70 loss remained substantially above the training loss, indicating the model's limited ability to extrapolate in terms of material composition. 

\begin{figure}[ht]
\centering
\includegraphics{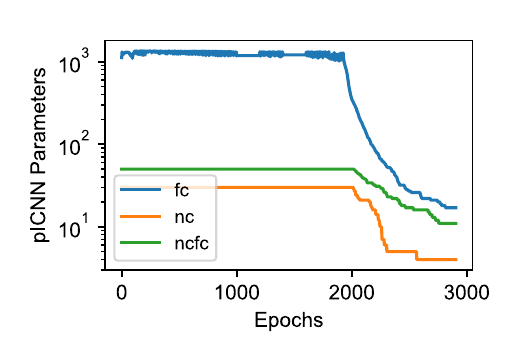}
\caption{Evolution of the total number of model parameters in the pICNN during training. Shown are the parameter counts for the fully-convex layers (fc), non-convex layers (nc), and the connection layers between the non-convex and fully-convex parts of the network (ncfc).}
\label{fig:num_parameters}
\end{figure}

The $L_{0}$ sparsification of the pICNN, performed in step 7, was intended to enhance the model's ability to predict unseen deformation conditions and material compositions while improving the interpretability of the final model. Figure \ref{fig:num_parameters} shows the parameter counts for the fully-convex layers (fc), non-convex layers (nc), and the connection layers between the non-convex and fully-convex parts of the network (ncfc). During training steps 1-6, the $L_{0}$-regularization terms were not applied ($\alpha_{fc}=\alpha_{nc}=\alpha_{ncfc}=0$). The fc layers had over 1000 non-zero parameters with small fluctuations during training, 25 parameters for nc, and 50 parameters for ncfc. In step 7, $\alpha_{fc}$ was ramped from 0 to $5\cdot10^{-4}$, and $\alpha_{nc}$ and $\alpha_{ncfc}$ was ramped from 0 to $10^{-6}$ over 1000 epochs. Following the onset of sparsification, a significant reduction of parameters occurred during the first 500 epochs, followed by a plateau in the second half of step 7. The final number of active parameters was 17 for the fc layers, 4 for the nc layers, and 11 for the connection (ncfc) layers. Notably, in the non-convex part of the network, the second layer was entirely pruned, leaving all active weights concentrated in the first layer.

After sparsification using $L_0$ regularization, the hyperelastic strain energy function $\Psi^{NN}$ modeled with the pICNN is given by Equation \ref{eqn:pICNN strain energy function}, where $a_1 = 6.84602$, $a_2 = 6.86996$, $a_3 = 6.88268$, $a_4 = 6.93876$, and $b_1 = 2.37126$, $b_2 = 2.1961$, $b_3 = 2.56167$, $b_4 = 1.9888$.

\begin{equation}
\begin{aligned}
\hat{\Psi}^{NN} = 
&\quad \Bigg\{0.083089 I_1
+ 4.01198 \log \Bigg[
\left(
\frac{
e^{-0.207897 I_1 + 0.135303 I_2}
}{
\left(
\prod_{k=1}^{4} \left( e^{a_k c} + 1 \right)^{b_k} + 1
\right)^{0.021823}
}
+ 1
\right)^{1.16891}
\\
&\quad \times
\left(
\left(
\prod_{k=1}^{4} \left( e^{a_k c} + 1 \right)^{b_k} + 1
\right)^{0.049792}
e^{0.228289 I_1 - 0.204157 I_2}
+ 1
\right)^{0.733048}
e^{0.033506 I_1}
+ 1
\Bigg]
\\
&\quad + 0.809351 \log\left( e^{0.009634 I_1} + 1 \right)
\Bigg\}
\label{eqn:pICNN strain energy function}
\end{aligned}
\end{equation}

Additionally, the mapping from material composition $c$ to the $\gamma$ parameter used in the quasi-linear viscoelastic model is given by Equation \ref{eqn:gamma_qlv}.

\begin{equation}
\begin{aligned}
\hat{\gamma} &= \left[
\frac{1}{
\displaystyle
\frac{p_1 p_2}{q_1} + 1
}
\right] \\
p_1 &= \left( 1 + e^{-1.381 c} \right)^{0.423}
\left( 1 + e^{-1.018 c} \right)^{0.446}
\left( 1 + e^{-0.976 c} \right)^{0.552}
\left( 1 + e^{-0.296 c} \right)^{0.492}, \\
p_2 &= \left( e^{0.059 c} + 1 \right)^{0.054}
\left( e^{0.111 c} + 1 \right)^{0.147}, \\
q_1 &= \left( e^{0.754 c} + 1 \right)^{0.254}
\left( e^{1.295 c} + 1 \right)^{0.264}.
\end{aligned}
\label{eqn:gamma_qlv}
\end{equation}

\begin{figure}[!ht]
\includegraphics[width=\textwidth]{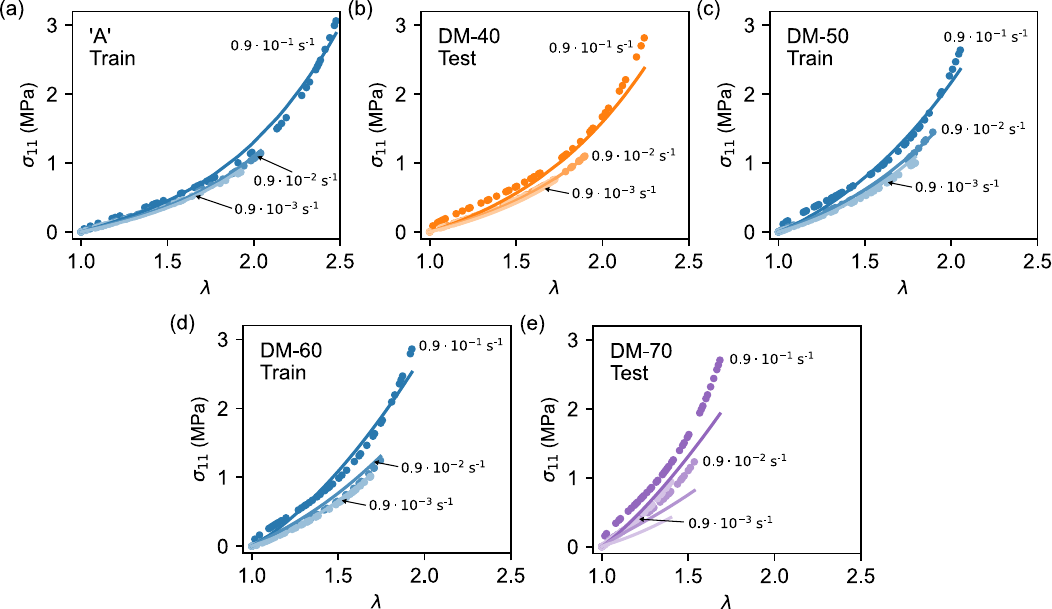}
\caption{Comparison of model predictions (lines) and experimental data (points) for uniaxial tension. Axial Cauchy stress $\sigma_{11}$ is plotted as a function of the stretch ratio $\lambda$. Experimental datasets A, DM-50, and DM-60 were used for model training, while DM-40 and DM-70 were used for testing.}
\label{fig:model_result_tension}
\end{figure}

\begin{figure}[!ht]
\includegraphics[width=\textwidth]{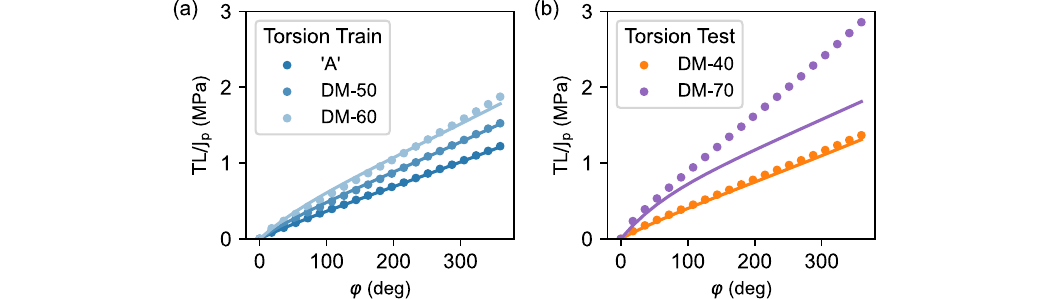}
\caption{Comparison of model prediction (lines) and experimental data (points) for torsion experiments conducted at a rotation rate of $360$ deg/min. $TL/J_{p}$ (MPa) versus rotation angle $\varphi$ (deg) is plotted for (a) experimental datasets used for training and (b) held out material compositions used for testing.}
\label{fig:model_result_torsion}
\end{figure}

Figures \ref{fig:model_result_tension} and \ref{fig:model_result_torsion} present the model prediction results for uniaxial tension and torsion, respectively. In Figure \ref{fig:model_result_tension}, axial Cauchy stress $\sigma_{11}$ is plotted against stretch ratio $\lambda$, while Figure \ref{fig:model_result_torsion} shows the normalized torque $TL/J_p$ as a function of the angle of twist. For the training compositions (A, DM-50, and DM-60), the model captures the overall shape of the response in both uniaxial tension and torsion. However, it shows small deviations in two regions of the uniaxial tension data: the small stress bump at low stretches and the sharp upturn at high stretches are both slightly underpredicted. These discrepancies likely arise from a combination of limited training data in those regions, the smoothing effect of model sparsification through aggressive $L_0$-regularization, and the constraint of fitting both tension and torsion data simultaneously. Together, these factors reduce the model's ability to represent localized nonlinear features at the extremes of the deformation range.

For the test data, the model predicts DM-40 well, consistent with its material composition falling within the training range. This is evident from the good fit across different strain rates in uniaxial tension, as well as the excellent torsion predictions for DM-40. In contrast, the prediction for DM-70, which lies outside the trained composition range, is poor, highlighting limited extrapolation performance for material composition. Specifically, the data at lower deformation rates (0.0009 $\text{s}^{-1}$ for tension, and $360$ deg/min for torsion) is poorly fitted for DM-70. This highlights the MLP network's extrapolation limitations in mapping material composition $c$ to the QLV model's $\gamma$ parameter. Moreover, since the material composition parameter passes through the PICNN's non-convex component, the higher $c$ value for DM-70 compared to DM-60 does not ensure reliable extrapolation, so the higher initial stiffness observed at the fastest rate is largely coincidental.

Table~\ref{tbl:metrics} provides a quantitative summary of the model's performance across different material compositions and loading conditions. For material compositions used during training (e.g., A, DM-50, DM-60), the model achieves consistently high accuracy. Out of 24 training and test datasets for these compositions, 21 have an $R^2$ above 0.98, and 22 have a sMAPE below 8\%. For held-out compositions within the training range, such as DM-40, predictive performance is strong with $R^2$ values exceed 0.96 and sMAPE typically remains under 14\%. However, for DM-70, whose composition lies outside the training range, model performance drops significantly. At the lowest tension rate (0.0009 $\text{s}^{-1}$), the $R^2$ value falls to 0.251 and sMAPE rises sharply to 34.5\%, indicating poor fit and limited generalization. The torsion prediction for DM-70 has a negative $R^2$ of -0.118 reflecting a breakdown in predictive reliability. These results highlight the model's sensitivity to extrapolation in material composition, especially at lower deformation rates where viscoelastic effects are more pronounced.

\begin{table*}[!ht]
\caption{Train and test evaluation metrics for various digital material composition input}
\begin{tabular}{llll|lll}
\hline
\multicolumn{4}{c|}{Tension}                                                                                                             & \multicolumn{3}{c}{Torsion}                                                                              \\ \hline
\multicolumn{1}{l|}{}            & \multicolumn{1}{c|}{rate (s$^{-1}$)} & \multicolumn{1}{c|}{$R^{2}$} & \multicolumn{1}{c|}{sMAPE (\%)} & \multicolumn{1}{c|}{rate ($^\circ$/min)} & \multicolumn{1}{c|}{$R^{2}$} & \multicolumn{1}{c}{sMAPE (\%)} \\ \hline
\multicolumn{1}{l|}{A train}     & \multicolumn{1}{l|}{0.09}            & \multicolumn{1}{l|}{0.992}   & 5.2                             & \multicolumn{1}{l|}{360}                 & \multicolumn{1}{l|}{0.999}   & 0.9                            \\
\multicolumn{1}{l|}{}            & \multicolumn{1}{l|}{0.009}           & \multicolumn{1}{l|}{0.996}   & 4.3                             & \multicolumn{1}{l|}{}                    & \multicolumn{1}{l|}{}        &                                \\
\multicolumn{1}{l|}{}            & \multicolumn{1}{l|}{0.0009}          & \multicolumn{1}{l|}{0.988}   & 6.8                             & \multicolumn{1}{l|}{}                    & \multicolumn{1}{l|}{}        &                                \\ \hline
\multicolumn{1}{l|}{A test}      & \multicolumn{1}{l|}{0.09}            & \multicolumn{1}{l|}{0.994}   & 11.7                            & \multicolumn{1}{l|}{360}                 & \multicolumn{1}{l|}{0.999}   & 0.6                            \\
\multicolumn{1}{l|}{}            & \multicolumn{1}{l|}{0.009}           & \multicolumn{1}{l|}{0.997}   & 3.5                             & \multicolumn{1}{l|}{}                    & \multicolumn{1}{l|}{}        &                                \\
\multicolumn{1}{l|}{}            & \multicolumn{1}{l|}{0.0009}          & \multicolumn{1}{l|}{0.995}   & 5.8                             & \multicolumn{1}{l|}{}                    & \multicolumn{1}{l|}{}        &                                \\ \hline
\multicolumn{1}{l|}{DM-40 test}  & \multicolumn{1}{l|}{0.09}            & \multicolumn{1}{l|}{0.964}   & 13.8                            & \multicolumn{1}{l|}{360}                 & \multicolumn{1}{l|}{0.990}   & 3.2                            \\
\multicolumn{1}{l|}{\textit{(held-out}}            & \multicolumn{1}{l|}{0.009}           & \multicolumn{1}{l|}{0.999}   & 2.0                             & \multicolumn{1}{l|}{}                    & \multicolumn{1}{l|}{}        &                                \\
\multicolumn{1}{l|}{\textit{composition)}}            & \multicolumn{1}{l|}{0.0009}          & \multicolumn{1}{l|}{0.971}   & 7.0                             & \multicolumn{1}{l|}{}                    & \multicolumn{1}{l|}{}        &                                \\ \hline
\multicolumn{1}{l|}{DM-50 train} & \multicolumn{1}{l|}{0.09}            & \multicolumn{1}{l|}{0.980}   & 7.8                             & \multicolumn{1}{l|}{360}                 & \multicolumn{1}{l|}{0.999}   & 1.3                            \\
\multicolumn{1}{l|}{}            & \multicolumn{1}{l|}{0.009}           & \multicolumn{1}{l|}{0.995}   & 6.4                             & \multicolumn{1}{l|}{}                    & \multicolumn{1}{l|}{}        &                                \\
\multicolumn{1}{l|}{}            & \multicolumn{1}{l|}{0.0009}          & \multicolumn{1}{l|}{0.985}   & 3.7                             & \multicolumn{1}{l|}{}                    & \multicolumn{1}{l|}{}        &                                \\ \hline
\multicolumn{1}{l|}{DM-50 test}  & \multicolumn{1}{l|}{0.09}            & \multicolumn{1}{l|}{0.980}   & 15.5                            & \multicolumn{1}{l|}{360}                 & \multicolumn{1}{l|}{0.999}   & 0.3                            \\
\multicolumn{1}{l|}{}            & \multicolumn{1}{l|}{0.009}           & \multicolumn{1}{l|}{0.991}   & 1.7                             & \multicolumn{1}{l|}{}                    & \multicolumn{1}{l|}{}        &                                \\
\multicolumn{1}{l|}{}            & \multicolumn{1}{l|}{0.0009}          & \multicolumn{1}{l|}{0.982}   & 2.9                             & \multicolumn{1}{l|}{}                    & \multicolumn{1}{l|}{}        &                                \\ \hline
\multicolumn{1}{l|}{DM-60 train} & \multicolumn{1}{l|}{0.09}            & \multicolumn{1}{l|}{0.977}   & 8.4                             & \multicolumn{1}{l|}{360}                 & \multicolumn{1}{l|}{0.994}   & 2.2                            \\
\multicolumn{1}{l|}{}            & \multicolumn{1}{l|}{0.009}           & \multicolumn{1}{l|}{0.956}   & 7.1                             & \multicolumn{1}{l|}{}                    & \multicolumn{1}{l|}{}        &                                \\
\multicolumn{1}{l|}{}            & \multicolumn{1}{l|}{0.0009}          & \multicolumn{1}{l|}{0.996}   & 5.1                             & \multicolumn{1}{l|}{}                    & \multicolumn{1}{l|}{}        &                                \\ \hline
\multicolumn{1}{l|}{DM-60 test}  & \multicolumn{1}{l|}{0.09}            & \multicolumn{1}{l|}{0.984}   & 9.9                             & \multicolumn{1}{l|}{360}                 & \multicolumn{1}{l|}{0.991}   & 1.9                            \\
\multicolumn{1}{l|}{}            & \multicolumn{1}{l|}{0.009}           & \multicolumn{1}{l|}{0.925}   & 7.6                             & \multicolumn{1}{l|}{}                    & \multicolumn{1}{l|}{}        &                                \\
\multicolumn{1}{l|}{}            & \multicolumn{1}{l|}{0.0009}          & \multicolumn{1}{l|}{0.994}   & 2.4                             & \multicolumn{1}{l|}{}                    & \multicolumn{1}{l|}{}        &                                \\ \hline
\multicolumn{1}{l|}{DM-70 test}  & \multicolumn{1}{l|}{0.09}            & \multicolumn{1}{l|}{0.798}   & 16.9                            & \multicolumn{1}{l|}{360}                 & \multicolumn{1}{l|}{-0.118}  & 15.5                           \\
\multicolumn{1}{l|}{\textit{(held-out}}            & \multicolumn{1}{l|}{0.009}           & \multicolumn{1}{l|}{0.751}   & 12.4                            & \multicolumn{1}{l|}{}                    & \multicolumn{1}{l|}{}        &                                \\
\multicolumn{1}{l|}{\textit{composition)}}            & \multicolumn{1}{l|}{0.0009}          & \multicolumn{1}{l|}{0.251}   & 34.5                            & \multicolumn{1}{l|}{}                    & \multicolumn{1}{l|}{}        &                                \\ \hline
\end{tabular}
\label{tbl:metrics}
\end{table*}

\section{Conclusions \label{sec:Conclusions}}

We presented a combined experimental and computational study of polymeric materials produced via Stratasys PolyJet multi-material 3D printing. These digital materials, composed of soft Agilus and rigid Digital ABS feedstocks, are microscale composites formed through droplet-by-droplet deposition during the PolyJet printing process. Fix distinct compositions (A, DM-40, DM-50, DM-60, and DM-70) were investigated to assess the influence of material composition on mechanical behavior. We conducted mechanical testing under both uniaxial tension and torsion across three deformation rates. These experiments revealed that non-linear and rate-dependent behavior varies strongly with the multi-material composition. Increasing Digital ABS content led to higher stiffness and reduced extensibility. Deformation rate effects were observed in the tension experiments, where a higher stretch rate of $\dot{\lambda} = 0.9\cdot10^{-1}\ \text{s}^{-1}$ resulted in increased stresses across all compositions.  In contrast, the lower rates of $\dot{\lambda} = 0.9\cdot10^{-3}\ \text{s}^{-1}$ produced lower, nearly identical stresses, indicating a transition to a rate-independent, quasi-static regime. The torsion experiments tested at $90$, $180$, and $360$ deg/min (1.6, 3.2, and 6.3 $\text{deg}/\text{min}\cdot \text{mm}$) had similar torque response, indicating that the fastest torsion rate remained within the quasi-static regime. Additionally, all material compositions exhibited Poisson's ratios around 0.47, justifying the assumption of incompressibility, a simplification retained in our physics-augmented constitutive model discovery process.

To model this complex mechanical behavior, we developed a physics-augmented machine learning framework that constructs a unified, composition-aware constitutive model trained directly on experimental data. The approach combines a partially input convex neural network to learn a convex hyperelastic strain energy function with a quasi-linear viscoelastic model, in which the relaxation parameter is predicted from material composition via an MLP neural network. This formulation eliminates the need to calibrate separate hyperelastic and viscoelastic models for each material variant and generalizes across both loading conditions and compositions.

The model achieved high accuracy across training data, with $R^2 > 0.98$ in 21 of 24 datasets and sMAPE below 8\% in 22 cases. It also interpolated well to unseen intermediate compositions (e.g., DM-40), though extrapolation beyond the trained composition range (e.g., DM-70) revealed limitations in generalizability particularly, due to non-convex dependencies on composition. To enhance interpretability and reduce model complexity, we applied $L_0$ sparsification, yielding a compact formulation with a minimal set of active parameters.

Looking forward, the PANN framework offers a scalable path toward data-driven modeling of multi-material systems. In this study, the time-dependent behavior was modeled using a fixed QLV structure, with only the relaxation coefficient learned from data. Future work, enabled by richer datasets with less noise, could extend this approach to discover full viscoelastic constitutive relations directly from data without relying on predefined forms. Additionally, this method could be expanded to polymer blends and composites with three or more components and user-defined micro-structures, further leveraging the capabilities of voxel-level material control in multi-material 3D printing.

\section*{Acknowledgements}
S.Y., N.B., R.E.J. and D.T.S. were supported by the Laboratory Directed Research and Development program at Sandia National Laboratories, a multimission laboratory managed and operated by National Technology and Engineering Solutions of Sandia, LLC, a wholly owned subsidiary of Honeywell International, Inc., for the U.S. Department of Energy's National Nuclear Security Administration under contract DE-NA-0003525. 
This paper describes objective technical results and analysis. Any subjective views or opinions that might be expressed in the paper do not necessarily represent the views of the U.S. Department of Energy or the United States Government. M.L. is a fellow of the Ariane de Rothschild Women Doctoral Program.

\providecommand{\latin}[1]{#1}
\makeatletter
\providecommand{\doi}
  {\begingroup\let\do\@makeother\dospecials
  \catcode`\{=1 \catcode`\}=2 \doi@aux}
\providecommand{\doi@aux}[1]{\endgroup\texttt{#1}}
\makeatother
\providecommand*\mcitethebibliography{\thebibliography}
\csname @ifundefined\endcsname{endmcitethebibliography}  {\let\endmcitethebibliography\endthebibliography}{}

\end{document}